\documentclass[12pt]{article}
\usepackage[a4paper, margin=2cm]{geometry}
\usepackage{amsmath, amssymb}
\usepackage{algorithm}
\usepackage{algpseudocode}

\usepackage[colorlinks=true, urlcolor=blue]{hyperref}
\usepackage{url}
\hypersetup{
    colorlinks=true,
    linkcolor=blue,
    citecolor=blue,
    urlcolor=blue
}

\usepackage[x11names, svgnames, rgb]{xcolor}
\usepackage{tikz}
\usetikzlibrary{arrows.meta, shapes.geometric, positioning, calc, arrows, snakes, shapes}
\usepackage{graphicx}
\usepackage{adjustbox}
\usepackage{wasysym}

\usepackage{listings}
\usepackage{tcolorbox}
\tcbuselibrary{listings,skins}

\definecolor{pythonorange}{RGB}{240,130,0}    
\definecolor{pythonblue}{RGB}{0,112,200}      	
\definecolor{pythongreen}{RGB}{64,128,80}    	
\definecolor{pythonpurple}{RGB}{170,34,255}   

\newtcblisting{pythoncode}[1][]{
    listing only,
    listing options={
        language=Python,
        basicstyle=\ttfamily\footnotesize, 
        columns=flexible,
        breaklines=true,
        showstringspaces=false,
        tabsize=4,
        keepspaces=true,
        showtabs=false,
        commentstyle=\color{pythongreen},
        keywordstyle=\color{pythonblue},
        stringstyle=\color{pythonpurple},
        emphstyle=\color{pythonorange},
        emph={super,__init__,__add__,__call__,__getitem__,__setitem__,__getattr__,__setattr__},
        morekeywords={self},
    },
    boxsep=2pt,        	
    left=3pt,          	
    right=3pt,         	
    top=3pt,           	
    bottom=3pt,        	
    arc=0mm,           	
    boxrule=0.5pt,     	
    colback=gray!5,    
    colframe=gray!50, 
    enhanced,
    #1
}

\DeclareMathOperator*{\argmin}{arg\,min}

\title{Introduction to Predictive Coding Networks for Machine Learning}
\author{Mikko Stenlund\footnote{\texttt{at pm dot me}}}

\date{May 29, 2025}

\begin{document}

\maketitle

\begin{abstract}
Predictive coding networks (PCNs) constitute a biologically inspired framework for understanding hierarchical computation in the brain, and offer an alternative to traditional feedforward neural networks in ML. This note serves as a quick, onboarding introduction to PCNs for machine learning practitioners. We cover the foundational network architecture, inference and learning update rules, and algorithmic implementation. A concrete image-classification task (CIFAR-10) is provided as a benchmark-smashing application, together with an accompanying Python notebook containing the PyTorch implementation.
\end{abstract}

\tableofcontents

\section{Introduction}
\label{sec:introduction}

The goal of this document is to present a first introduction to predictive coding networks from both conceptual and algorithmic standpoints, in the context of machine learning. We focus on detailing the structure of PCNs, deriving their inference and learning rules, and demonstrating their usefulness in an experiment. This section provides a brief overview to set the stage.

Predictive coding is a foundational theory in neuroscience proposing that the brain is fundamentally a prediction machine, constantly attempting to anticipate incoming sensory inputs and updating internal representations to minimize the difference between expectations and actual observations. This concept forms the basis of the predictive coding network (PCN), a class of hierarchical generative models designed to mirror these principles in artificial systems.

Already in 1867, Helmholtz proposed that perception is an unconscious inferential process where the brain predicts how planned actions will affect sensory inputs \cite{Helmholtz1867}. In the mid-20th century, Barlow’s efficient coding hypothesis posited that the brain economizes neural representation by removing predictable redundancies in sensory signals \cite{Barlow1961}, foreshadowing a later focus on unexpected or surprising sensory events. Gregory further argued that perception is a constructive, hypothesis-driven endeavor, with visual illusions highlighting how top-down expectations shape perception \cite{Gregory1980}. By the 1990s, theoretical models explicitly invoked hierarchical prediction: Mumford proposed a cortical architecture in which higher-level areas send predictions to lower-level areas, with only residual errors fed forward \cite{Mumford1992}. Such concepts set the stage for the predictive coding model of visual cortex by Rao and Ballard \cite{rao1999predictive}, which formalized perception as a hierarchical interplay of top-down predictions and bottom-up error signals.

Originally developed to explain extra-classical receptive field effects in visual cortex \cite{rao1999predictive}, predictive coding has since been generalized into a unified framework for cortical processing through the \textit{free-energy principle} \cite{friston2005theory, friston2010free}. This principle casts perception and action as inference problems, where organisms minimize a variational bound on surprise or prediction error.

Neurophysiological plausibility of predictive coding has been further explored in works such as \cite{spratling2008reconciling}, which unify predictive coding with biased competition models of attention, and \cite{bastos2012canonical}, which describe potential microcircuit implementations in cortical hierarchies. Keller and Mrsic-Flogel \cite{keller2018predictive} offer evidence that predictive processing is a canonical computation across the cortex, with supportive anatomical and functional data reviewed by Shipp \cite{shipp2016neural}. Further empirical neuroscience evidence is presented in \cite{Walsh2020,Caucheteux2023}.

From a computational modeling perspective, predictive coding networks provide an alternative to traditional feedforward models trained via backpropagation. Notably, Whittington and Bogacz \cite{whittington2017approximation} demonstrated that predictive coding networks can approximate backpropagation in multilayer neural networks using only local updates; see also Millidge et al.~\cite{millidge2022predictive}. Their implementation in spiking neural networks was reviewed in \cite{ndri2024predictive}.

Predictive coding has also inspired innovations in unsupervised and self-supervised learning. For example, Schmidhuber \cite{Schmidhuber1992Predictability} proposed the predictability minimization principle echoing the redundancy reduction goal of predictive coding. Lotter et al.~\cite{lotter2017deep} introduced deep predictive coding networks for video frame prediction, offering state-of-the-art performance using unsupervised objectives. Moreover, \cite{HaSchmidhuber2018} developed a “world model”---a network that learns a latent predictive representation of an agent’s environment---showing how hierarchical prediction and the minimization of surprise can facilitate efficient learning and planning.

A broader theoretical and empirical review of predictive coding in both neuroscience and artificial intelligence is provided in \cite{millidge2020predictive}, which also outlines future research directions.

The references cited here are just a bite-sized sample of the enormous literature on the subject. The interested reader is advised to look into the bibliographies in those for more coverage, and to search the arXiv for the latest developments. There are also many excellent blog posts on the subject, such as \cite{millidge2020blog, alonso2022blog}.

\section{Network Architecture}

\paragraph{Model.}
A PCN consists of $L\ge 1$ layers of latent variables \( \mathbf{x}^{(l)} \in \mathbb{R}^{d_l}\), \(1\le l\le L\), and an input layer of variables \( \mathbf{x}^{(0)} \in \mathbb{R}^{d_0}\). Each layer attempts to predict the state of the layer {\bf below}, so the architecture includes the following {\bf top-down} elements for \(0\le l\le L-1\):
\begin{itemize}
	\item Weights \( \mathbf{W}^{(l)} \in\mathbb{R}^{d_l\times d_{l+1}}\) from layer \(l+1\) to layer \(l\)
	\item Preactivations
	\[
	\mathbf{a}^{(l)} = \mathbf{W}^{(l)} \mathbf{x}^{(l+1)} \in \mathbb{R}^{d_l}
	\]
	\item Predictions
	\[
	\hat{\mathbf{x}}^{(l)} = f^{(l)}(\mathbf{a}^{(l)}) \in \mathbb{R}^{d_l}
	\]
	where $f^{(l)}$ is, often a nonlinear, scalar function applied elementwise
	\item Prediction errors
	\[ 
	\boldsymbol{\varepsilon}^{(l)} = \mathbf{x}^{(l)} - \hat{\mathbf{x}}^{(l)} \in \mathbb{R}^{d_l}
	\]
\end{itemize}
The loss function subject to minimization is the total square prediction error, or {\bf energy}:
\[
\mathcal{L} = \frac{1}{2} \sum_{l=0}^{L-1}  {\bigl\| \boldsymbol{\varepsilon}^{(l)} \bigr\|^2} \ .
\]

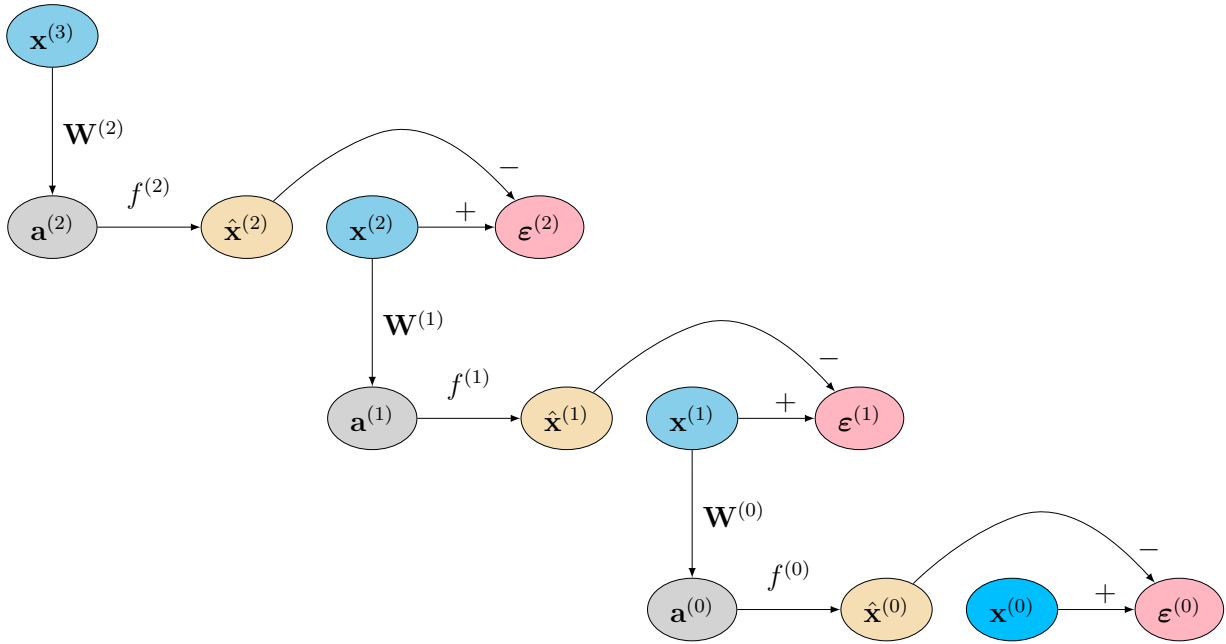
\begin{figure}[htbp]
\centering
\begin{adjustbox}{width=0.95\linewidth}
	\begin{tikzpicture}[>=latex,line join=bevel,]
\begin{scope}
  \pgfsetstrokecolor{black}
  \definecolor{strokecol}{rgb}{1.0,1.0,1.0};
  \pgfsetstrokecolor{strokecol}
  \definecolor{fillcol}{rgb}{1.0,1.0,1.0};
  \pgfsetfillcolor{fillcol}
  \filldraw (0.0bp,0.0bp) -- (0.0bp,263.2bp) -- (490.92bp,263.2bp) -- (490.92bp,0.0bp) -- cycle;
\end{scope}
\begin{scope}
  \pgfsetstrokecolor{black}
  \definecolor{strokecol}{rgb}{1.0,1.0,1.0};
  \pgfsetstrokecolor{strokecol}
  \definecolor{fillcol}{rgb}{1.0,1.0,1.0};
  \pgfsetfillcolor{fillcol}
  \filldraw (0.0bp,0.0bp) -- (0.0bp,263.2bp) -- (490.92bp,263.2bp) -- (490.92bp,0.0bp) -- cycle;
\end{scope}
\begin{scope}
  \pgfsetstrokecolor{black}
  \definecolor{strokecol}{rgb}{1.0,1.0,1.0};
  \pgfsetstrokecolor{strokecol}
  \definecolor{fillcol}{rgb}{1.0,1.0,1.0};
  \pgfsetfillcolor{fillcol}
  \filldraw (0.0bp,0.0bp) -- (0.0bp,263.2bp) -- (490.92bp,263.2bp) -- (490.92bp,0.0bp) -- cycle;
\end{scope}
  \node (preact_2) at (16.734bp,173.75bp) [draw,fill=LightGray,ellipse] {$\mathbf{a}^{(2)}$};
  \node (pred_2) at (95.734bp,173.75bp) [draw,fill=Wheat,ellipse] {$\hat{\mathbf{x}}^{(2)}$};
  \node (error_2) at (214.73bp,173.75bp) [draw,fill=LightPink,ellipse] {$\boldsymbol{\varepsilon}^{(2)}$};
  \node (lat_2) at (146.73bp,173.75bp) [draw,fill=SkyBlue,ellipse] {$\mathbf{x}^{(2)}$};
  \node (preact_1) at (146.73bp,95.45bp) [draw,fill=LightGray,ellipse] {$\mathbf{a}^{(1)}$};
  \node (pred_1) at (225.73bp,95.45bp) [draw,fill=Wheat,ellipse] {$\hat{\mathbf{x}}^{(1)}$};
  \node (error_1) at (344.73bp,95.45bp) [draw,fill=LightPink,ellipse] {$\boldsymbol{\varepsilon}^{(1)}$};
  \node (lat_1) at (276.73bp,95.45bp) [draw,fill=SkyBlue,ellipse] {$\mathbf{x}^{(1)}$};
  \node (preact_0) at (276.73bp,17.15bp) [draw,fill=LightGray,ellipse] {$\mathbf{a}^{(0)}$};
  \node (pred_0) at (355.73bp,17.15bp) [draw,fill=Wheat,ellipse] {$\hat{\mathbf{x}}^{(0)}$};
  \node (error_0) at (474.73bp,17.15bp) [draw,fill=LightPink,ellipse] {$\boldsymbol{\varepsilon}^{(0)}$};
  \node (inp_0) at (406.73bp,17.15bp) [draw,fill=DeepSkyBlue,ellipse] {$\mathbf{x}^{(0)}$};
  \node (lat_3) at (16.734bp,252.05bp) [draw,fill=SkyBlue,ellipse] {$\mathbf{x}^{(3)}$};
  \draw [->] (lat_3) ..controls (16.734bp,229.17bp) and (16.734bp,211.02bp)  .. (preact_2);
  \definecolor{strokecol}{rgb}{0.0,0.0,0.0};
  \pgfsetstrokecolor{strokecol}
  \draw (33.734bp,212.9bp) node {$\mathbf{W}^{(2)}$};
  \draw (10.734bp,234.44bp) node {$$};
  \draw (10.734bp,191.27bp) node {$$};
  \draw [->] (preact_2) ..controls (43.371bp,173.75bp) and (56.136bp,173.75bp)  .. (pred_2);
  \draw (56.065bp,187.75bp) node {$f^{(2)}$};
  \draw (39.488bp,167.75bp) node {$$};
  \draw (72.732bp,167.75bp) node {$$};
  \draw [->] (pred_2) ..controls (115.05bp,192.72bp) and (131.95bp,206.67bp)  .. (149.73bp,212.38bp) .. controls (168.01bp,218.24bp) and (186.62bp,204.8bp)  .. (error_2);
  \draw (155.73bp,212.9bp) node {$$};
  \draw (119.32bp,187.82bp) node {$$};
  \draw (202.35bp,198.15bp) node {$-$};
  \draw [->] (lat_2) ..controls (170.9bp,173.75bp) and (179.17bp,173.75bp)  .. (error_2);
  \draw (181.01bp,182.75bp) node {$$};
  \draw (177.49bp,167.41bp) node {$$};
  \draw (184.69bp,180.09bp) node {$+$};
  \draw [->] (lat_2) ..controls (146.73bp,150.87bp) and (146.73bp,132.72bp)  .. (preact_1);
  \draw (163.73bp,134.6bp) node {$\mathbf{W}^{(1)}$};
  \draw (140.73bp,156.14bp) node {$$};
  \draw (140.73bp,112.97bp) node {$$};
  \draw [->] (preact_1) ..controls (173.37bp,95.45bp) and (186.14bp,95.45bp)  .. (pred_1);
  \draw (186.07bp,109.45bp) node {$f^{(1)}$};
  \draw (169.49bp,89.45bp) node {$$};
  \draw (202.73bp,89.45bp) node {$$};
  \draw [->] (pred_1) ..controls (245.05bp,114.42bp) and (261.95bp,128.37bp)  .. (279.73bp,134.08bp) .. controls (298.01bp,139.94bp) and (316.62bp,126.5bp)  .. (error_1);
  \draw (285.73bp,134.6bp) node {$$};
  \draw (249.32bp,109.52bp) node {$$};
  \draw (332.35bp,119.85bp) node {$-$};
  \draw [->] (lat_1) ..controls (300.9bp,95.45bp) and (309.17bp,95.45bp)  .. (error_1);
  \draw (311.01bp,104.45bp) node {$$};
  \draw (307.49bp,89.111bp) node {$$};
  \draw (314.69bp,101.79bp) node {$+$};
  \draw [->] (lat_1) ..controls (276.73bp,72.566bp) and (276.73bp,54.42bp)  .. (preact_0);
  \draw (293.73bp,56.3bp) node {$\mathbf{W}^{(0)}$};
  \draw (270.73bp,77.841bp) node {$$};
  \draw (270.73bp,34.669bp) node {$$};
  \draw [->] (preact_0) ..controls (303.37bp,17.15bp) and (316.14bp,17.15bp)  .. (pred_0);
  \draw (316.07bp,31.15bp) node {$f^{(0)}$};
  \draw (299.49bp,11.15bp) node {$$};
  \draw (332.73bp,11.15bp) node {$$};
  \draw [->] (pred_0) ..controls (375.05bp,36.119bp) and (391.95bp,50.071bp)  .. (409.73bp,55.775bp) .. controls (428.01bp,61.64bp) and (446.62bp,48.198bp)  .. (error_0);
  \draw (415.73bp,56.3bp) node {$$};
  \draw (379.32bp,31.219bp) node {$$};
  \draw (462.35bp,41.545bp) node {$-$};
  \draw [->] (inp_0) ..controls (430.9bp,17.15bp) and (439.17bp,17.15bp)  .. (error_0);
  \draw (441.01bp,26.15bp) node {$$};
  \draw (437.49bp,10.811bp) node {$$};
  \draw (444.69bp,23.489bp) node {$+$};
\end{tikzpicture}
\end{adjustbox}
\caption{A graph representation of a PCN with three latent layers.}
\label{fig:pcn}
\end{figure}

The network can be viewed as a directed acyclic graph; see Figure~\ref{fig:pcn}. The ``hanging'' root nodes without incoming edges are the input and latent variables; the leaves (end nodes) are the prediction errors. Observe that the generative hierarchy flows from top to bottom, \emph{towards} the input; more on this will be discussed shortly. 

\vspace{1em}

\paragraph{Alternating minimization procedure.}
The generative weights \( \mathbf{W} = ( \mathbf{W}^{(0)},\dots,\mathbf{W}^{(L-1)} ) \) define the overall shape of the predictive coding energy landscape. For fixed weights and input, the {\bf inference} process seeks a configuration \( \mathbf{x}^* \) of latent values \( \mathbf{x} = ( \mathbf{x}^{(1)},\dots,\mathbf{x}^{(L)} ) \) that minimizes the energy:
\[
\mathbf{x}^* = \argmin_{\mathbf{x}} \mathcal{L}(\mathbf{x}; \mathbf{W}, \mathbf{x}^{(0)})\ .
\]
Optimization then switches to {\bf learning}, which seeks to perturb the energy landscape by adjusting the weights gently to \(\mathbf{W}' = \mathbf{W} + \delta\mathbf{W}\) (e.g., taking one gradient step with a small learning rate) so as to further reduce the energy at the inferred configuration \( \mathbf{x}^* \):
\[
\mathcal{L}(\mathbf{x}^*; \mathbf{W}', \mathbf{x}^{(0)}) < \mathcal{L}(\mathbf{x}^*; \mathbf{W}, \mathbf{x}^{(0)}) \ .
\]
In this way, inference performs descent within a fixed landscape, while learning deforms the landscape to better accommodate future inference. This separation of roles gives predictive coding networks their characteristic two-timescale dynamics: fast inference, slow learning.

In practice, the procedure repeats for a new input \(\mathbf{x}^{(0)}\) and reinitialized latent configuration~\(\mathbf{x}\). Such matters are deferred to the last section where an actual application will be discussed together with the training details. 

\paragraph{Convergence.}
Several works have established, formally and experimentally, that the alternating optimization procedure in predictive coding networks—consisting of inference steps over latent variables followed by learning updates of the generative weights—converges to a local minimum of the loss under sufficient assumptions \cite{whittington2017approximation, song2020can, frieder2022non,mali2024tight,alonso2022theoretical,salvatori2024stable}. That said, what matters to practitioners is performance in task-specific settings rather than just guarantees of local convergence under idealized conditions. On that note, the application in the last section serves as one example where fast convergence and excellent generalization do occur.

\paragraph{Generative hierarchy.}
The network is built in stacked layers of latent variables. The idea is that each layer represents the data at a different level of abstraction, from raw sensory inputs up through progressively higher‐order features. Predictions flow top-down (higher layers predicting lower‐layer activity) and errors flow bottom-up. Similar processes are believed to exist in the brain’s multi‐level sensory hierarchies. Rather than just learning a discriminative mapping from inputs to labels, the PCN explicitly encodes a generative model of how lower‐level activity could be produced from higher‐level causes, so the network can attempt to reconstruct what the sensory data should look like given its latent beliefs.

\paragraph{A word about hybrid predictive coding.}
In the algorithms and implementation presented in this document, the initial values of the latent variables are random---both literally and figuratively. Recently, an interesting hybrid predictive coding model has been proposed, where the latents~\(\mathbf{x}^{(l)}\) are initialized to the values \(\boldsymbol{\xi}^{(l)}\) predicted by another network \cite{tschantz2022hybrid}. The predictions of this second network flow in the bottom-up direction, opposite to the PCN hierarchy, via ``amortized'' functions \(\mathbf{g}^{(l)}:\mathbb{R}^{d_{l-1}}\to \mathbb{R}^{d_{l}}\) in a feedforward fashion: \(\boldsymbol{\xi}^{(l)}= \mathbf{g}^{(l)}(\boldsymbol{\xi}^{(l-1)})\), \(1\le l\le L\), starting from the input \(\boldsymbol{\xi}^{(0)} = \mathbf{x}^{(0)}\). Such a construction reflects the adjustment of the network's posterior beliefs upon receiving sensory input, before the refining inference process of the vanilla PCN begins. The rest of this note does not involve the hybrid model.

\section{Inference and Learning Rules}

\subsection{Latent state update rule (inference)}

We aim to update each latent variable \( \mathbf{x}^{(l)} \) (\( 1\le l\le L\)) via gradient descent on \( \mathcal{L} \).

\noindent\underline{\bf Case \(1\le l<L\).} The variable \( \mathbf{x}^{(l)} \) appears in two places in the loss:
in \( \boldsymbol{\varepsilon}^{(l)} = \mathbf{x}^{(l)} -  \hat{\mathbf{x}}^{(l)}\) explicitly and
in \( \boldsymbol{\varepsilon}^{(l-1)} = \mathbf{x}^{(l-1)} - \hat{\mathbf{x}}^{(l-1)} \) through \( \hat{\mathbf{x}}^{(l-1)} =  f^{(l-1)}(\mathbf{W}^{(l-1)} \mathbf{x}^{(l)}) \) .
Hence,
\[
\frac{\partial \mathcal{L}}{\partial x_i^{(l)}} 
= 
\sum_j \varepsilon_j^{(l)} \frac{\partial \varepsilon_j^{(l)}}{\partial x_i^{(l)}} +
\sum_j \varepsilon_j^{(l-1)} \frac{\partial \varepsilon_j^{(l-1)}}{\partial x_i^{(l)}} 
= 
\varepsilon_i^{(l)} -
\sum_j \varepsilon_j^{(l-1)} \frac{\partial \hat x_j^{(l-1)}}{\partial x_i^{(l)}} 
\]
where
\[
\frac{\partial \hat x_j^{(l-1)}}{\partial x_i^{(l)}} 
=
\frac{\partial}{\partial x_i^{(l)}} (f^{(l-1)}(a_{j}^{(l-1)}))
=
f^{(l-1)'}(a_{j}^{(l-1)})W_{ji}^{(l-1)}\ .
\]
Hence, the gradient with respect to  \( \mathbf{x}^{(l)} \) is
\[
\nabla_{\mathbf{x}^{(l)}}\mathcal{L}
=
\boldsymbol{\varepsilon}^{(l)}
- \mathbf{W}^{(l-1)\top} \left( f^{(l-1)'}(\mathbf{a}^{(l-1)}) \odot \boldsymbol{\varepsilon}^{(l-1)} \right)
\qquad (1\le l< L)
\]
where  \( \odot \) denotes the elementwise (Hadamard) product, $f^{(l-1)'}$ is the derivative of $f^{(l-1)}$, and the convention used for stacking the entries is that the vectors \( \mathbf{x}^{(l)} \) and \( \nabla_{\mathbf{x}^{(l)}}\mathcal{L} \) have the same shape.

\noindent\underline{\bf Case \(l=L\).} Notice that \( \mathbf{x}^{(L)} \) only appears inside \( \boldsymbol{\varepsilon}^{(L-1)} \) in the expression of \(\mathcal{L}\). Hence,
\[
\frac{\partial \mathcal{L}}{\partial x_i^{(L)}} 
= 
\sum_j \varepsilon_j^{(L-1)} \frac{\partial \varepsilon_j^{(L-1)}}{\partial x_i^{(L)}}
=
-\sum_j \varepsilon_j^{(L-1)} \frac{\partial \hat x_j^{(L-1)}}{\partial x_i^{(L)}} 
=
-\sum_j \varepsilon_j^{(L-1)} f^{(L-1)'}(a_{j}^{(L-1)})W_{ji}^{(L-1)}\ .
\]
In vector form,
\[
\nabla_{\mathbf{x}^{(L)}}\mathcal{L}
=
- \mathbf{W}^{(L-1)\top} \left( f^{(L-1)'}(\mathbf{a}^{(L-1)}) \odot \boldsymbol{\varepsilon}^{(L-1)} \right)
\]
which is similar to \(1\le l < L\) except for the missing first term: there is no prediction error for the top layer.

Introducing the convenient constant
\[
\boldsymbol{\varepsilon}^{(L)} = \boldsymbol{0}
\]
the {\bf inference update} rule for the gradient descent algorithm becomes compactly
\[
\boxed{
\mathbf{x}^{(l)} \leftarrow \mathbf{x}^{(l)} - \eta_{\text{infer}} \left( \boldsymbol{\varepsilon}^{(l)} - \mathbf{W}^{(l-1)\top} \left( f^{(l-1)'}(\mathbf{a}^{(l-1)}) \odot \boldsymbol{\varepsilon}^{(l-1)} \right) \right)
} \qquad (1\le l\le L)
\]
where $\eta_{\text{infer}}>0$ is an inference rate of choice.

During inference, all prediction errors and feedback terms are computed first using the current network state, and only then are the latent variables \( \mathbf{x}^{(l)} \) updated. This ensures that each update step is based on a consistent energy landscape and avoids using partially updated states within the same iteration. Conceptually, this corresponds to a synchronous update scheme where all neurons compute their next state based on the same network snapshot.

\subsection{Weight update rule (learning)}

Each weight matrix \( \mathbf{W}^{(l)} \) is responsible for predicting \( \mathbf{x}^{(l)} \) from \( \mathbf{x}^{(l+1)} \), and appears only in
\(
\boldsymbol{\varepsilon}^{(l)} = \mathbf{x}^{(l)} - f^{(l)}(\mathbf{W}^{(l)} \mathbf{x}^{(l+1)})
\).
To minimize the loss, we compute
\[
\frac{\partial \mathcal{L}}{\partial W_{ij}^{(l)}} = \sum_k \varepsilon_k^{(l)} \frac{\partial \varepsilon_k^{(l)}}{\partial W_{ij}^{(l)}} 
= - \sum_k \varepsilon_k^{(l)} \frac{\partial}{\partial W_{ij}^{(l)}} (f^{(l)} (a_{k}^{(l)}))
\]
which yields
\[
\frac{\partial \mathcal{L}}{\partial W_{ij}^{(l)}} 
= - \sum_k \varepsilon_k^{(l)} f^{(l)'}(a_{k}^{(l)}) \delta_{ik}  x_j^{(l+1)}
= - \varepsilon_i^{(l)} f^{(l)'}(a_{i}^{(l)})  x_j^{(l+1)} \ .
\]
As before, using the convention that the matrices \(\mathbf{W}^{(l)}\) and \( \nabla_{\mathbf{W}^{(l)}}\mathcal{L} \) have the same shape, we arrive at the gradient
\[
\nabla_{\mathbf{W}^{(l)}}\mathcal{L}
=
- \left( f^{(l)'}(\mathbf{a}^{(l)}) \odot \boldsymbol{\varepsilon}^{(l)} \right) \mathbf{x}^{(l+1)\top} \ .
\]

In particular, the {\bf learning update} rule via gradient descent is
\[
\boxed{
\mathbf{W}^{(l)} \leftarrow \mathbf{W}^{(l)} + \eta_{\text{learn}} \left(f^{(l)'}(\mathbf{a}^{(l)} ) \odot \boldsymbol{\varepsilon}^{(l)}\right) \mathbf{x}^{(l+1)\top}
}  \qquad (0\le l< L)
\]
with a learning rate of $\eta_{\text{learn}}>0$.

Notice that the quantities
\[
\mathbf{h}^{(l)} = f^{(l)'}( \mathbf{a}^{(l)} ) \odot \boldsymbol{\varepsilon}^{(l)} \qquad (0\le l<L)
\]
are central, as they appear in the expressions of the gradients with respect to both the latents and the weights. They could be called {\bf gain-modulated errors}; see the subsection below.

\subsection{Locality of the updates}
One of the original motivations behind predictive coding is its potential biological plausibility: that the brain could implement something akin to deep hierarchical learning using local computations. 
Locality typically refers to whether a computation depends only on information from a given layer and its immediate neighbors. This concept is important both for computational efficiency and biological plausibility.

The \textbf{learning update} for the weight matrix \( \mathbf{W}^{(l)} \) is \textbf{local in a strong sense}: it depends only on the local activity \( \mathbf{x}^{(l+1)} \) of layer \( l+1 \) (presynaptic) and the local prediction error \( \boldsymbol{\varepsilon}^{(l)} \) at layer \( l \) (postsynaptic), making it compatible with Hebbian-like plasticity mechanisms often summarized as ``neurons that fire together, wire together.'' 

In contrast, while the \textbf{inference update} for a latent variable \( \mathbf{x}^{(l)} \)  depends only on adjacent layers, it requires access to the error signal \( \boldsymbol{\varepsilon}^{(l-1)} \) broadcast from the lower layer, modulated by the top-down weights in \( \mathbf{W}^{(l-1)} \). Thus, inference updates are layer-local but not neuron-local, since a neuron's update depends on a weighted sum of errors from other neurons in the layer below.

Biologically, the learning updates
\(
\delta W^{(l)}_{ij} = -\eta_{\text{learn}}  \varepsilon^{(l)}_i  f^{(l)'}(a^{(l)}_i) x^{(l+1)}_j
\)
are thought to be more plausible than inference updates.
Here, \( \varepsilon^{(l)}_i \) is the prediction error at the postsynaptic neuron~\( i \),
\(
a^{(l)}_i = \sum_m W^{(l)}_{im} x^{(l+1)}_m
\)
is its preactivation, and \( x^{(l+1)}_j \) is the activity of the presynaptic neuron~\( j \). The preactivation
represents the total synaptic input to neuron \( i \)—essentially its membrane potential or driving current. In both biological and artificial neurons, this summation is performed naturally as part of neural activation. The neuron does not need to access other neurons’ states; it simply integrates the inputs it receives via its dendrites.
Thus, \( f^{(l)'}(a^{(l)}_i) \) can be interpreted as a local gain or nonlinearity applied to the neuron's own internal state. This makes the full weight update neuron-local in the strong sense: it depends only on information accessible to the synapse between neurons \( j \to i \), including presynaptic activity, postsynaptic error, and internal quantities of the postsynaptic neuron.

\subsection{Motivations beyond biological plausibility}

While predictive coding networks (PCNs) are often motivated by neuroscience and cortical modeling, their architectural design makes them attractive for future machine learning systems on emerging (e.g., neuromorphic) hardware.

\paragraph{Locality and parallelism.}
PCNs rely on local computations: each weight update depends only on the activity and prediction error of adjacent neurons. This makes them well suited to distributed and parallel computing, in contrast to  global gradient backpropagation.

\paragraph{Separation of algorithmic and physical synchrony.}
Inference in PCNs is described here as a synchronous algorithm: each update step operates on a fixed snapshot of the network state. This ensures convergence and simplifies analysis. In principle, the underlying computations could be implemented asynchronously in hardware, without needing a global clock or centralized scheduling.

\paragraph{Decentralized control.}
Unlike backpropagation, which requires tightly coordinated forward and backward passes, PCNs do not rely on a global gradient tape or synchronized dataflow. This allows them to operate on hardware with minimal coordination or shared memory.

\paragraph{Energy-efficient, adaptive inference.}
PCNs support variable-length inference: predictable inputs can settle in fewer steps, while uncertain or surprising stimuli drive deeper inference. This adaptivity enables anytime computation, useful in energy-constrained settings such as embedded devices or robotics.

\paragraph{Architectural versatility.}
PCNs extend naturally to convolutional, recurrent, and graph-based structures by redefining local prediction and feedback pathways.

\vspace{1em}
In short, predictive coding offers a biologically inspired view of computation whose design pattern aligns with the demands of emerging hardware. These properties make PCNs a high-value target for research in scalable, low-power, and distributed learning systems \cite{davies2018loihi, millidge2022predictive}.

\section{Base Algorithms}
\subsection{Unsupervised learning in PCNs}

\noindent
This algorithm implements unsupervised learning in a predictive coding network with \( L\) layers of latent variables \( \mathbf{x}^{(1)}, \dots, \mathbf{x}^{(L)} \), with the input variables \( \mathbf{x}^{(0)} \) clamped to the input data; recall Figure~\ref{fig:pcn} in Section~\ref{sec:introduction}. The latent variables are inferred via iterative updates to minimize the global prediction error energy as discussed earlier, starting from a random initial state. Each inference step uses a consistent snapshot of the network: all prediction errors and gradients are computed before any latent state is updated.

Each layer \( l \) receives top-down predictions from layer \( l+1 \) through a learned weight matrix~\( \mathbf{W}^{(l)} \) and nonlinearity \( f^{(l)} \). After inference, weights are updated using local Hebbian-like learning rules based on the final prediction errors.

\paragraph*{Per-sample training.}
The base algorithm presented here describes a single inference-learning cycle for one input sample. Training proceeds by repeating this cycle over many samples drawn from a dataset. For each sample, latent variables are inferred and weights are updated once. For mini-batch training discussed later, the inference loop can be run for the samples in the current batch in parallel, and the subsequent weight update(s) can be carried out using the mean gradient over the batch.

\vspace{1em}
\begin{algorithm}[H]
\caption{Unsupervised learning in a predictive coding network}
\begin{algorithmic}[1]
\Require Input \( \mathbf{x}^{(0)} \), generative weights \( \{ \mathbf{W}^{(l)} \}_{l=0}^{L-1} \), activation functions and their derivatives \( \{ f^{(l)}, f^{(l)'} \} \), number of inference steps \( T_{\text{infer}} \), learning rate \( \eta_{\text{learn}} \), inference rate \( \eta_{\text{infer}} \)

\State \textbf{Clamp} \( \mathbf{x}^{(0)} \gets \text{input data} \)
\For{layer \( l = 1 \) to \( L \)} 
    \State Initialize \( \mathbf{x}^{(l)} \gets \text{small random values} \)
\EndFor
\State \( \boldsymbol{\varepsilon}^{(L)} \gets \mathbf{0} \) \Comment{Top layer has no prediction error}

\For{step \( t = 1 \) to \( T_{\text{infer}} \)} \Comment{{\bf Inference update loop}}
	\For{layer \( l = 0 \) to \( L-1 \)} \Comment{Store network state snapshot}
	    \State \( \mathbf{a}^{(l)} \gets \mathbf{W}^{(l)} \mathbf{x}^{(l+1)} \) \Comment{Preactivation}
	    \State \( \hat{\mathbf{x}}^{(l)} \gets f^{(l)}(\mathbf{a}^{(l)}) \) \Comment{Prediction}
	    \State \( \boldsymbol{\varepsilon}^{(l)} \gets \mathbf{x}^{(l)} - \hat{\mathbf{x}}^{(l)} \) \Comment{Prediction error}
	\EndFor

    \For{layer \( l = 1 \) to \( L \)} \Comment{Update latents using snapshot}
        \State \( \mathbf{g}_{\mathbf{x}}^{(l)} \gets \boldsymbol{\varepsilon}^{(l)} - \mathbf{W}^{(l-1)\top} \left( f^{(l-1)'}(\mathbf{a}^{(l-1)}) \odot \boldsymbol{\varepsilon}^{(l-1)} \right) \) \Comment{Gradient wrt $\mathbf{x}^{(l)}$}
        \State \( \mathbf{x}^{(l)} \gets \mathbf{x}^{(l)} - \eta_{\text{infer}} \, \mathbf{g}_{\mathbf{x}}^{(l)} \)
    \EndFor
\EndFor

\For{layer \( l = 0 \) to \( L-1 \)} \Comment{{\bf Weight update}}
    \State \( \mathbf{a}^{(l)} \gets \mathbf{W}^{(l)} \mathbf{x}^{(l+1)} \)
    \State \( \hat{\mathbf{x}}^{(l)} \gets f^{(l)}(\mathbf{a}^{(l)}) \)
    \State \( \boldsymbol{\varepsilon}^{(l)} \gets \mathbf{x}^{(l)} - \hat{\mathbf{x}}^{(l)} \)
    \State \( \mathbf{g}_{\mathbf{W}}^{(l)} \gets - \left( \boldsymbol{\varepsilon}^{(l)} \odot f^{(l)'}(\mathbf{a}^{(l)}) \right) \mathbf{x}^{(l+1)\top} \) \Comment{Gradient wrt $\mathbf{W}^{(l)}$}
    \State \( \mathbf{W}^{(l)} \gets \mathbf{W}^{(l)} - \eta_{\text{learn}} \, \mathbf{g}_{\mathbf{W}}^{(l)}\)
\EndFor
\end{algorithmic}
\end{algorithm}

\paragraph{Remark.} The model can also support anytime inference, where well-predicted inputs converge in fewer steps, and additional inference steps are taken to improve predictions in ambiguous cases. This adaptivity offers potential energy savings in embedded or neuromorphic deployments. The algorithm can be modified in this spirit by choosing a sufficiently large maximum step count \(T_{\text{infer}}\), and running the inference loop until either \(T_{\text{infer}}\) steps have been performed or convergence has been detected. Here convergence  means, for instance, that the norm of the latest update (or updates over a longer patience window) across all latent variables falls below a preset threshold. In machine learning terminology, this could be phrased as inference with sample-wise early stopping.

\subsection{Supervised learning extension}

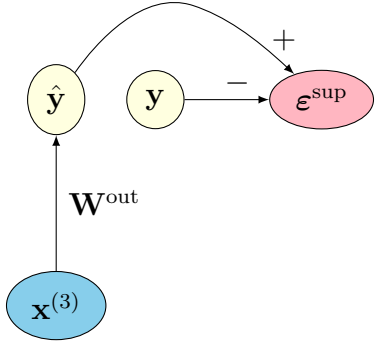
\begin{figure}[htbp]
\centering
	\begin{tikzpicture}[>=latex,line join=bevel,]
\begin{scope}
  \pgfsetstrokecolor{black}
  \definecolor{strokecol}{rgb}{1.0,1.0,1.0};
  \pgfsetstrokecolor{strokecol}
  \definecolor{fillcol}{rgb}{1.0,1.0,1.0};
  \pgfsetfillcolor{fillcol}
  \filldraw (0.0bp,0.0bp) -- (0.0bp,141.06bp) -- (133.16bp,141.06bp) -- (133.16bp,0.0bp) -- cycle;
\end{scope}
  \node (label_y) at (53.734bp,94.555bp) [draw,fill=LightYellow,ellipse] {$\mathbf{y}$};
  \node (pred_y) at (16.734bp,94.555bp) [draw,fill=LightYellow,ellipse] {$\hat{\mathbf{y}}$};
  \node (error_sup) at (115.73bp,94.555bp) [draw,fill=LightPink,ellipse] {$\boldsymbol{\varepsilon}^{\rm sup}$};
  \node (lat_3) at (16.734bp,17.15bp) [draw,fill=SkyBlue,ellipse] {$\mathbf{x}^{(3)}$};
  \draw [->] (label_y) ..controls (69.672bp,94.555bp) and (78.162bp,94.555bp)  .. (error_sup);
  \definecolor{strokecol}{rgb}{0.0,0.0,0.0};
  \pgfsetstrokecolor{strokecol}
  \draw (80.662bp,103.56bp) node {$$};
  \draw (77.016bp,88.216bp) node {$$};
  \draw (84.29bp,100.89bp) node {$-$};
  \draw [<-] (pred_y) ..controls (16.734bp,57.577bp) and (16.734bp,39.652bp)  .. (lat_3);
  \draw (34.734bp,55.8bp) node {$\mathbf{W}^{\mathrm{out}}$};
  \draw (10.734bp,77.07bp) node {$$};
  \draw (10.734bp,34.546bp) node {$$};
  \draw [->] (pred_y) ..controls (31.58bp,116.04bp) and (49.307bp,135.99bp)  .. (67.734bp,129.93bp) .. controls (79.434bp,126.09bp) and (90.626bp,118.18bp)  .. (error_sup);
  \draw (61.734bp,129.81bp) node {$$};
  \draw (35.902bp,110.74bp) node {$$};
  \draw (101.77bp,117.1bp) node {$+$};
\end{tikzpicture}
\caption{A supervised extension of the PCN with three latent layers in Figure~\ref{fig:pcn} is obtained by stacking a readout layer on top of the highest latent layer, together with a new root node for the target label.}
\label{fig:pcn_sup}
\end{figure}

A minimal modification to apply predictive coding in a supervised setting entails simply clamping the top latent representation \( \mathbf{x}^{(L)} \) to a predicted label \( \hat{\mathbf{y}} \in \mathbb{R}^{d_{\text{out}}} \), treating it as part of the generative hierarchy.
For the sake of variation in this pedagogically oriented note, we introduce a separate readout layer that maps \( \mathbf{x}^{(L)} \mapsto \hat{\mathbf{y}} \) linearly:
\[
\hat{\mathbf{y}} = \mathbf{W}^{\text{out}} \mathbf{x}^{(L)}
\]
where \( \mathbf{W}^{\text{out}}\in\mathbb{R}^{d_{\text{out}}\times d_L} \). Given a target label \( \mathbf{y}  \in \mathbb{R}^{d_{\text{out}}} \), we define a supervised error
\[
\boldsymbol{\varepsilon}^{\text{sup}} = \hat{\mathbf{y}} - \mathbf{y} \ .
\]
Figure~\ref{fig:pcn_sup} illustrates these changes relative to Figure~\ref{fig:pcn}. The loss function (energy) now becomes
\[
\mathcal{L} + \mathcal{L}_{\text{sup}}
\]
where \(
\mathcal{L}_{\text{sup}} = \frac{1}{2} \|\boldsymbol{\varepsilon}^{\text{sup}} \|^2
\)
is the supervised energy.

\vspace{1em}

The supervised error is backpropagated into the top latent representation \( \mathbf{x}^{(L)} \) during inference. This does not change the update rules of the lower latents \( \mathbf{x}^{(l)} \) or the generative weights~\(\mathbf{W}^{(l)}\), \(0\le l<L\), at all.
Since $\nabla_{\mathbf{x}^{(L)}} \mathcal{L}_{\text{sup}} = \mathbf{W}^{\text{out}\top} \boldsymbol{\varepsilon}^{\text{sup}}
$, the inference update rule for the top latent is modified to 
\[
\boxed{
\mathbf{x}^{(L)} \leftarrow \mathbf{x}^{(L)} - \eta_{\text{infer}} \left(\mathbf{W}^{\text{out}\top} \boldsymbol{\varepsilon}^{\text{sup}}- \mathbf{W}^{(L-1)\top} \left( f^{(L-1)'}(\mathbf{a}^{(L-1)}) \odot \boldsymbol{\varepsilon}^{(L-1)} \right) \right) 
} \ .
\]
Note that this becomes formally the same as in the unsupervised case if we introduce the symbol \( \boldsymbol{\varepsilon}^{(L)} = \mathbf{W}^{\text{out}\top} \boldsymbol{\varepsilon}^{\text{sup}}\) (instead of \( \boldsymbol{\varepsilon}^{(L)} = \boldsymbol{0} \)), which is how we implement the algorithm.

On the other hand,
\(
\nabla_{\mathbf{W}^{\text{out}}} \mathcal{L}_{\text{sup}} = \boldsymbol{\varepsilon}^{\text{sup}} \mathbf{x}^{(L)\top} .
\)
The output weights \( \mathbf{W}^{\text{out}} \) are thus updated via
\[
\boxed{
\mathbf{W}^{\text{out}} \gets \mathbf{W}^{\text{out}} - \eta_{\text{learn}} \, \boldsymbol{\varepsilon}^{\text{sup}} \mathbf{x}^{(L)\top} 
}
\ .
\]

The core network structure remains unchanged from the unsupervised case; the only modification is the supervised error signal applied to the top layer.

\vspace{1em}
\begin{algorithm}[H]
\caption{Supervised learning in a predictive coding network}
\begin{algorithmic}[1]
\Require Input \( \mathbf{x}^{(0)} \), target label \( \mathbf{y} \), generative weights \( \{ \mathbf{W}^{(l)} \}_{l=0}^{L-1} \), output weights \( \mathbf{W}^{\text{out}} \), activation functions and their derivatives \( \{ f^{(l)}, f^{(l)'} \} \), inference steps \( T_{\text{infer}} \), learning rate \( \eta_{\text{learn}} \), inference rate \( \eta_{\text{infer}} \)

\State \textbf{Clamp} \( \mathbf{x}^{(0)} \gets \text{input data} \)
\For{layer \( l = 1 \) to \( L \)} 
    \State Initialize \( \mathbf{x}^{(l)} \gets \text{small random values} \)
\EndFor

\For{step \( t = 1 \) to \( T_{\text{infer}} \)} \Comment{{\bf Inference update loop}}
    \For{layer \( l = 0 \) to \( L-1 \)}  \Comment{Store network state snapshot}
        \State \( \mathbf{a}^{(l)} \gets \mathbf{W}^{(l)} \mathbf{x}^{(l+1)} \) \Comment{Preactivation}
        \State \( \hat{\mathbf{x}}^{(l)} \gets f^{(l)}(\mathbf{a}^{(l)}) \) \Comment{Prediction}
        \State \( \boldsymbol{\varepsilon}^{(l)} \gets \mathbf{x}^{(l)} - \hat{\mathbf{x}}^{(l)} \) \Comment{Prediction error}
    \EndFor
    \State \( \hat{\mathbf{y}} \gets \mathbf{W}^{\text{out}} \mathbf{x}^{(L)} \) \Comment{Output}
    \State \( \boldsymbol{\varepsilon}^{\text{sup}} \gets \hat{\mathbf{y}} - \mathbf{y} \) \Comment{Supervised error}
    \State \( \boldsymbol{\varepsilon}^{(L)} \gets \mathbf{W}^{\text{out}\top} \boldsymbol{\varepsilon}^{\text{sup}}\) \Comment{Convenient notation}

    \For{layer \( l = 1 \) to \( L \)} \Comment{Update latents using snapshot}
        \State \( \mathbf{g}_\mathrm{x}^{(l)} \gets \boldsymbol{\varepsilon}^{(l)} - \mathbf{W}^{(l-1)\top} \left( f^{(l-1)'}(\mathbf{a}^{(l-1)}) \odot \boldsymbol{\varepsilon}^{(l-1)} \right) \)
        \State \( \mathbf{x}^{(l)} \gets \mathbf{x}^{(l)} - \eta_{\text{infer}} \, \mathbf{g}_\mathrm{x}^{(l)} \)
    \EndFor
\EndFor

\For{layer \( l = 0 \) to \( L-1 \)} \Comment{{\bf Weight update}}
    \State \( \mathbf{a}^{(l)} \gets \mathbf{W}^{(l)} \mathbf{x}^{(l+1)} \)
    \State \( \hat{\mathbf{x}}^{(l)} \gets f^{(l)}(\mathbf{a}^{(l)}) \)
    \State \( \boldsymbol{\varepsilon}^{(l)} \gets \mathbf{x}^{(l)} - \hat{\mathbf{x}}^{(l)} \)
    \State \( \mathbf{g}_{\mathbf{W}}^{(l)} \gets - \left( \boldsymbol{\varepsilon}^{(l)} \odot f^{(l)'}(\mathbf{a}^{(l)}) \right) \mathbf{x}^{(l+1)\top} \)
    \State \( \mathbf{W}^{(l)} \gets \mathbf{W}^{(l)} - \eta_{\text{learn}} \,  \mathbf{g}_{\mathbf{W}}^{(l)} \)
\EndFor

\State \( \mathbf{g}_\mathbf{W}^{\mathrm{out}} \gets \boldsymbol{\varepsilon}^{\text{sup}} \mathbf{x}^{(L)\top}\) \Comment{Gradient wrt $\mathbf{W}^{\mathrm{out}}$}
\State \( \mathbf{W}^{\text{out}} \gets \mathbf{W}^{\text{out}} - \eta_{\text{learn}} \, \mathbf{g}_\mathbf{W}^{\mathrm{out}}  \)
\end{algorithmic}
\end{algorithm}

\section{Application: Supervised Learning on CIFAR-10}

To demonstrate the practical applicability of predictive coding networks, we evaluate a supervised PCN on the CIFAR-10 image classification task. CIFAR-10 consists of 60{,}000 color images of size \(32 \times 32 \times 3\), divided evenly across 10 classes, with 50{,}000 training and 10{,}000 test samples~\cite{krizhevsky2009learning}.

\subsection{Model architecture}

Each input image is flattened into a vector \( \mathbf{x}^{(0)} \in \mathbb{R}^{3072} \), normalized to the range \([0,1]\). The network is defined with \(L = 3\) latent layers, where the topmost latent state \( \mathbf{x}^{(L)} \in \mathbb{R}^{10} \) is linearly mapped to an output vector \( \hat{\mathbf{y}} \in \mathbb{R}^{10} \) using a readout matrix \( \mathbf{W}^{\text{out}} \in \mathbb{R}^{10 \times 10} \).
The latent representations are
\[
\mathbf{x}^{(1)} \in \mathbb{R}^{1000}, \quad
\mathbf{x}^{(2)} \in \mathbb{R}^{500}, \quad
\mathbf{x}^{(3)} \in \mathbb{R}^{10}
\]
and the top-down generative weights are
\[
\mathbf{W}^{(0)} \in \mathbb{R}^{3072 \times 1000}, \quad 
\mathbf{W}^{(1)} \in \mathbb{R}^{1000 \times 500}, \quad 
\mathbf{W}^{(2)} \in \mathbb{R}^{500 \times 10}
\]
with scalar nonlinearity \( f^{(l)} = \mathrm{ReLU} = \max(0,\,\cdot\,)\) applied elementwise. No bias terms are used. The network structure is quite arbitrary, following just the battle-tested practice that the layer dimensions interpolate between the input and output in a quick progression.

The total number of trainable parameters in the PCN is
\[
 3{,}072 \times 1{,}000 + 1{,}000 \times 500 + 500 \times 10 + 10 \times 10 = 3{,}577{,}100\ .
\]
To draw a very loose connection with biology, the PCN comprises roughly \(4.6\times 10^3\) ``neurons'' (3{,}072 inputs, 1{,}000 first-order latents, 500 second-order latents, and 10 outputs) connected by \(3.6\times 10^6\) ``synapses.''

\subsection{Training procedure}
All weights are initialized using Xavier initialization, once, at the very beginning.

Each input-label pair \( (\mathbf{x}^{(0)}, y) \) undergoes the inference and learning cycle, as described in the supervised learning algorithm, with the latent variables \( \mathbf{x}^{(1)}, \dots, \mathbf{x}^{(L)} \) freshly initialized to small Gaussian noise. 
The output prediction is computed as
\(
\hat{\mathbf{y}} = \mathbf{W}^{\text{out}} \mathbf{x}^{(L)}
\)
and the supervised prediction error is given by
\(
\boldsymbol{\varepsilon}^{\text{sup}} = \hat{\mathbf{y}} - \mathbf{y}
\)
where \( \mathbf{y} \in \mathbb{R}^{10} \) is the one-hot encoded target \(y\).

Although the base algorithm was formulated in a per-sample manner, mini-batch training improves computational efficiency and learning stability. We process a batch of \( B \) samples \( \{ (\mathbf{x}^{(0)}_b, \mathbf{y}_b) \}_{b=1}^B \) in parallel. Each sample maintains its own set of latent variables \( \mathbf{x}^{(l)}_b \) and errors~\( \boldsymbol{\varepsilon}^{(l)}_b \), updated independently using the usual inference rule over \( T \) iterations.
After inference, the weight gradients $\mathbf{g}_b^{(l)} = -( \boldsymbol{\varepsilon}^{(l)}_b \odot f^{(l)'}(\mathbf{a}^{(l)}_b) ) \mathbf{x}^{(l+1)\top}_b$ and $\mathbf{g}_b^{\text{out}} =  \boldsymbol{\varepsilon}^{\text{sup}}_b \mathbf{x}^{(L)\top}_b$ are computed per sample (as in the base algorithm), and averaged over the batch:
\[
\mathbf{W}^{(l)} \gets \mathbf{W}^{(l)} - \eta_{\text{learn}} \frac{1}{B} \sum_{b=1}^B \mathbf{g}_b^{(l)}
\]
and
\[
\mathbf{W}^{\text{out}} \gets  \mathbf{W}^{\text{out}}  - \eta_{\text{learn}} \frac{1}{B} \sum_{b=1}^B \mathbf{g}_b^{\text{out}}
\]
Since the total number of inference updates per batch is \(B\times T_{\text{infer}}\), these learning updates are performed \(B\) times, thus keeping the ratio of inference updates to learning updates at \(T_{\text{infer}}\). For each learning step the same inputs \( \mathbf{x}^{(0)}_b \) and inferred latents \( \mathbf{x}^{(l)}_b \) are used to recompute the gradients \(\mathbf{g}_b^{(l)}\) and \( \mathbf{g}_b^{\text{out}} \).

This procedure preserves the sample-wise locality of PCN inference and learning, while leveraging parallel processing on a GPU.

\paragraph{Hyperparameters.} For training we set the hyperparameters as follows:
\begin{itemize}
    \item Batch size: 500 (resulting in 100 train batches, 20 test batches)
    \item Inference steps per sample: \( T_{\text{infer}} = 50 \)
    \item Inference rate: \( \eta_{\text{infer}} = 0.05 \)
    \item Learning steps per batch: \( T_{\text{learn}} = 500 \)  (equals batch size)
    \item Learning rate: \( \eta_{\text{learn}} = 0.005 \)
\end{itemize}
Such choices are largely guided by the belief, or inductive bias, that there should be a distinct {\bf separation of timescales}: inference should progress much faster than learning.

\subsection{Testing}
To test the trained network, the weights \(\mathbf{W}^{(l)}\) and \(\mathbf{W}^{\mathrm{out}}\) are frozen. For each input-label pair \( (\mathbf{x}^{(0)}, y) \) from the test set, the latents \(\mathbf{x}^{(l)}\) are initialized randomly and the inference loop is executed, exactly as in the base algorithm. Once the latents are optimized, the prediction \( \hat{\mathbf{y}} \) is read from the output layer. Observe that, given an input \( \mathbf{x}^{(0)} \), the prediction \( \hat{\mathbf{y}} \) in practice contains some random noise, since the inference loop starts from randomly initialized latents.

Top-1 and top-3 class-prediction accuracies are used as performance metrics: Let \(\mathrm{top}_k(\hat{\mathbf{y}})\) be the set of \( k \) indices corresponding to the largest entries of \(\hat{\mathbf{y}}\).  Then the top-\(k\) accuracy is the average of \(\mathbf{1}(y \in \mathrm{top}_k(\hat{\mathbf{y}}))\) over the test set. Note that the test accuracies for a well-trained PCN vary slightly from one test round to the next, due to the random initialization.

In the implementation test samples are handled in batches, just like in training.

\subsection{PyTorch implementation}

It is customary in PyTorch to load data as tensors where the batch dimension comes first. In our case the shape of a batch of samples is then \((B, d_0)\), corresponding to a matrix \(\mathbf{X}^{(0)}\) whose {\bf rows} represent the inputs \(\mathbf{x}_b^{(0)}\in \mathbb{R}^{1\times d_0} \), \(1\le b\le B\). To adhere to the custom and to allow for efficient batch handling via vectorization, we revise the supervised learning algorithm to its final form. The changes from earlier essentially amount to adding a batch dimension to all vectors and transposing certain products. As explained above, the learning step is furthermore repeated \(B\) times (by default) instead of just once; this counter-balances the increase in the number of inference steps due to batching prior to any weight updates---from \(T_{\text{infer}}\) per sample in case of no batching (\(B=1\)) to \(B\times T_{\text{infer}}\) per batch. To make reasoning about dimensions easier, they are listed explicitly on the right side of the algorithm box.

During training we also track the batch-averaged total energy
\[
\mathcal{E}_\textrm{batch}(t) = \frac{1}{B} \sum_{b=1}^{B}\biggl(\frac12 \sum_{l=0}^{L-1}   {\| \boldsymbol{\varepsilon}_b^{(l)}(t) \|^2} +  \frac12 \|\boldsymbol{\varepsilon}_b^{\text{sup}}(t) \|^2 \biggr)
\]
at every inference and learning step~$t$. Here the quadratic terms are the per-sample energies. Since there are $100$ batches per epoch, we thus generate $100$ ``energy trajectories'' over the time window \((0,\dots,T_\mathrm{infer} + T_\mathrm{learn})\) per epoch. 
Note that for sufficiently small inference and learning rates these trajectories should be strictly decreasing through inference and learning, which can be used as a sanity check that the implementation is correct. In practice, larger rates are required to escape local minima and to facilitate faster convergence.

To support energy‐tracking, our implementation reorders a few lines compared to the pseudocode here. Specifically, we compute and log the errors 
\(\boldsymbol{\varepsilon}_b^{(l)}(t)\) and \(\boldsymbol{\varepsilon}_b^{\mathrm{sup}}(t)\) pre‐ and post‐update at each step, then reuse those same quantities for the next inference or weight update. This ensures that we record the energy both before any updates (at \(t=0\)) and immediately after each subsequent update, without redundant recomputation.

\vspace{1em}
\begin{algorithm}[H] 
\caption{Supervised learning in a PCN (vectorized row‐batch form)}
\begin{algorithmic}[1]
\Require 
  Input batch $\mathbf{X}^{(0)}\in\mathbb{R}^{B\times d_0}$, 
  target batch $\mathbf{Y}\in\mathbb{R}^{B\times d_{\mathrm{out}}}$,
  generative weights $\{\mathbf{W}^{(l)}\in\mathbb{R}^{d_l\times d_{l+1}}\}_{l=0}^{L-1}$, 
  output weights $\mathbf{W}^{\mathrm{out}}\in\mathbb{R}^{d_{\mathrm{out}}\times d_{L}}$,
  activation functions and their derivatives $\{f^{(l)},\,f^{(l)'}\}$, inference steps per sample $T_{\text{infer}}$, learning steps per batch $T_{\text{learn}} = B$, inference rate $\eta_{\mathrm{infer}}$, learning rate $\eta_{\mathrm{learn}}$
\For{layer \( l = 1 \) to \( L \)} 
    \State Initialize \( \mathbf{X}^{(l)} \gets \text{small random values} \) \Comment{$B\times d_l$}
\EndFor
\For{$t=1$ \textbf{to} $T_{\text{infer}}$}  \Comment{{\bf Inference update loop}}
  \For{$l=0$ \textbf{to} $L-1$}
    \State $\mathbf{A}^{(l)} \gets \mathbf{X}^{(l+1)}\,\mathbf{W}^{(l)\top}$ \Comment{$B\times d_l$ } 
    \State $\hat{\mathbf{X}}^{(l)} \gets f^{(l)}(\mathbf{A}^{(l)})$ \Comment{$B\times d_l$ }  
    \State $\mathbf{E}^{(l)} \gets \mathbf{X}^{(l)} - \hat{\mathbf{X}}^{(l)}$ \Comment{$B\times d_l$ } 
    \State $\mathbf{H}^{(l)} = \mathbf{E}^{(l)} \,\odot\, f^{(l)'}(\mathbf{A}^{(l)})$ \Comment{$B\times d_l$}
  \EndFor
  \State $\hat{\mathbf{Y}} \gets \mathbf{X}^{(L)}\,\mathbf{W}^{\mathrm{out}\top}$ \Comment{$B\times d_{\mathrm{out}}$}  
  \State $\mathbf{E}^{\mathrm{sup}} \gets \hat{\mathbf{Y}} - \mathbf{Y}  $ \Comment{$B\times d_{\mathrm{out}}$}  
  \State $\mathbf{E}^{(L)} \gets \mathbf{E}^{\mathrm{sup}}\;\mathbf{W}^{\mathrm{out}}$  \Comment{$B\times d_{L}$}  
  \For{$l=1$ \textbf{to} $L$}
    \State $\mathbf{G}_\mathbf{X}^{(l)} \gets \mathbf{E}^{(l)} - \mathbf{H}^{(l-1)}\,\mathbf{W}^{(l-1)}$  \Comment{Grads wrt sample latents $\mathbf{x}_b^{(l)}$, batched \qquad $\triangleright$  $B\times d_l$ } 
    \State $\mathbf{X}^{(l)} \gets \mathbf{X}^{(l)} - \eta_{\mathrm{infer}} \mathbf{G}_\mathbf{X}^{(l)}$ \Comment{$B\times d_l$ } 
  \EndFor
\EndFor

\For{$t=1$ \textbf{to} $T_{\text{learn}}$}  \Comment{{\bf Weight update loop}}
  \For{$l=0$ \textbf{to} $L-1$}
    \State $\mathbf{A}^{(l)} \gets \mathbf{X}^{(l+1)}\,\mathbf{W}^{(l)\top}$   \Comment{$B\times d_l$ }
    \State $\hat{\mathbf{X}}^{(l)} \gets f^{(l)}(\mathbf{A}^{(l)})$   \Comment{$B\times d_l$ }
    \State $\mathbf{E}^{(l)} \gets \mathbf{X}^{(l)} - \hat{\mathbf{X}}^{(l)}$   \Comment{$B\times d_l$ }
    \State $\mathbf{H}^{(l)} = \mathbf{E}^{(l)} \,\odot\, f^{(l)'}(\mathbf{A}^{(l)})$ \Comment{$B\times d_l$}
    \State $\mathbf{G}_\mathbf{W}^{(l)} \gets - \frac 1B \mathbf{H}^{(l)\top} \mathbf{X}^{(l+1)}$  \Comment{Batch-averaged grad wrt $\mathbf{W}^{(l)}$ \qquad $\triangleright$ $d_l\times d_{l+1}$}
    \State $\mathbf{W}^{(l)} \gets \mathbf{W}^{(l)} - \eta_{\mathrm{learn}}\,\mathbf{G}_\mathbf{W}^{(l)}$ \Comment{$d_l\times d_{l+1}$}
  \EndFor
    \State $\hat{\mathbf{Y}} \gets \mathbf{X}^{(L)}\,\mathbf{W}^{\mathrm{out}\top}$ \Comment{$B\times d_{\mathrm{out}}$}  
  \State $\mathbf{E}^{\mathrm{sup}} \gets \hat{\mathbf{Y}} - \mathbf{Y}  $ \Comment{$B\times d_{\mathrm{out}}$}  
  \State $\displaystyle \mathbf{G}_\mathbf{W}^{\mathrm{out}} \gets \frac 1B \mathbf{E}^{\mathrm{sup}\,\top}\,\mathbf{X}^{(L)}$ \Comment{Batch-averaged grad wrt $\mathbf{W}^{\mathrm{out}}$ \qquad $\triangleright$ $d_{\mathrm{out}}\times d_{L}$}
  \State $\mathbf{W}^{\mathrm{out}} \gets \mathbf{W}^{\mathrm{out}} - \eta_{\mathrm{learn}}\,\mathbf{G}_\mathbf{W}^{\mathrm{out}}$  \Comment{$d_{\mathrm{out}}\times d_{L}$}
\EndFor
\end{algorithmic}
\end{algorithm}

Next, key modules of the Python code will be described. Full implementation details are provided in the accompanying Python notebook, available in a \href{https://github.com/Monadillo/pcn-intro}{GitHub repository}~\cite{monadillo}.

\subsubsection{The \texttt{PCNLayer} class}
\label{sec:pcnlayer}

Each latent layer of the PCN is encapsulated by the \texttt{PCNLayer} class, defined as follows:

\begin{pythoncode}[title=PCNLayer class]
class PCNLayer(nn.Module):
    def __init__(self,
                 in_dim,
                 out_dim,
                 activation_fn=torch.relu,
                 activation_deriv=lambda a: (a > 0).float()
                 ):
        super().__init__()
        self.W = nn.Parameter(torch.empty(out_dim, in_dim))
        nn.init.xavier_uniform_(self.W)
        self.activation_fn     = activation_fn
        self.activation_deriv  = activation_deriv

    def forward(self, x_above):
        with autocast(device_type='cuda'):
            a     = x_above @ self.W.T
            x_hat = self.activation_fn(a)
            return x_hat, a
\end{pythoncode}

The \texttt{PCNLayer} class inherits from \texttt{torch.nn.Module}, making it a standard building block in the PyTorch ecosystem. Its constructor, \texttt{\_\_init\_\_}, initializes the following components:
\begin{itemize}
    \item \textbf{Weights (\texttt{self.W})}: The learnable generative weight matrix, corresponding to $\mathbf{W}^{(l)}$ in our model, is initialized as a \texttt{torch.nn.Parameter}. This ensures that the weights are recognized by PyTorch's autograd system and can be updated during the learning phase. The dimensions are \texttt{(out\_dim, in\_dim)}, representing $d_l \times d_{l+1}$, connecting layer $l+1$ (of \texttt{in\_dim} neurons) to layer $l$ (of \texttt{out\_dim} neurons) in a top-down predictive manner. Xavier uniform initialization (\texttt{nn.init.xavier\_uniform\_}) is used to set the initial values of these weights. No biases are used.
    \item \textbf{Activation function (\texttt{self.activation\_fn})}: This stores the element-wise nonlinear activation function $f^{(l)}$, which defaults to ReLU (\texttt{torch.relu}).
    \item \textbf{Activation derivative (\texttt{self.activation\_deriv})}: This stores the derivative of the activation function, $f^{(l)'}$, required for calculating the gain-modulated errors $\mathbf{H}^{(l)}$ during both inference and learning updates. The default is the derivative of ReLU.
\end{itemize}

The core computation of a layer---generating a prediction for the layer below---is handled by the \texttt{forward} method. Given an input \texttt{x\_above} (representing the state of the layer above, $\mathbf{X}^{(l+1)}$ in batch form), this method performs two main operations:
\begin{enumerate}
    \item It computes the pre-activation $\mathbf{A}^{(l)} = \mathbf{X}^{(l+1)}\mathbf{W}^{(l)\top}$ (denoted \texttt{a} in the code). This corresponds to the matrix multiplication \texttt{x\_above @ self.W.T}.
    \item It then applies the layer's activation function $f^{(l)}$ to the pre-activations \texttt{a} to produce the prediction $\hat{\mathbf{X}}^{(l)}$ (denoted \texttt{x\_hat} in the code).
\end{enumerate}
The method returns both the prediction \texttt{x\_hat} and the pre-activation \texttt{a}, as both are needed for subsequent error calculations and gradient computations as detailed in the vectorized algorithm (Algorithm~3). By encapsulating these operations in a single module, \texttt{PCNLayer} separates the per‐layer computations required for both inference (updating latents) and learning (updating weights) in Algorithm 3. The use of \texttt{autocast(device\_type='cuda')} is for mixed-precision training, which can improve computational efficiency on compatible hardware.

\subsubsection{The \texttt{PredictiveCodingNetwork} class}
\label{sec:pcn}
Building upon the \texttt{PCNLayer}, the \texttt{PredictiveCodingNetwork} class organizes the entire network hierarchy and its operations. This class, also inheriting from \texttt{torch.nn.Module}, manages the collection of layers, the readout mechanism for supervised tasks, and utility functions for initialization and error computation.

\begin{pythoncode}[title=PredictiveCodingNetwork class]
class PredictiveCodingNetwork(nn.Module):
    def __init__(self,
                 dims,
                 output_dim
                 ):
        super().__init__()
        self.dims = dims
        self.L = len(dims) - 1
        self.layers = nn.ModuleList([
            PCNLayer(in_dim=dims[l+1],
                     out_dim=dims[l])
            for l in range(self.L)
        ])
        self.readout = nn.Linear(dims[-1], output_dim, bias=False)

    def init_latents(self, batch_size, device):
        return [
           torch.randn(batch_size, d, device=device, 
           				requires_grad=False)
           for d in self.dims[1:]
        ]

    def compute_errors(self, inputs_latents):
        errors, gain_modulated_errors = [], []
        for l, layer in enumerate(self.layers):
            x_hat, a  = layer(inputs_latents[l + 1])
            err       = inputs_latents[l] - x_hat
            gm_err    = err * layer.activation_deriv(a)
            errors.append(err)
            gain_modulated_errors.append(gm_err)
        return errors, gain_modulated_errors
\end{pythoncode}

The \texttt{\_\_init\_\_} constructor takes
\[
\texttt{dims} = [d_0, d_1, \dots, d_L]
\quad\text{and}\quad
\texttt{output\_dim} = d_{\mathrm{out}}
\]
specifying the dimensions of each layer in the network (from the input's $d_0$ to the topmost latent's $d_L$ and readout's $d_{\mathrm{out}}$) and performs the following:
\begin{itemize}
    \item \textbf{Dimensions (\texttt{self.dims}, \texttt{self.L})}: The \texttt{dims} list and $\texttt{self.L = len(dims)-1} = L$ are stored.
    \item \textbf{Generative layers (\texttt{self.layers})}: A \texttt{torch.nn.ModuleList} is created to hold the stack of \texttt{PCNLayer} instances representing the $L$ latent layers. Each \texttt{PCNLayer} maps \(\mathbb{R}^{d_{l+1}}\to\mathbb{R}^{d_l}\) as in the generative model. The dimensions are drawn from \texttt{dims}.
    \item \textbf{Readout layer (\texttt{self.readout})}: For supervised learning, a \texttt{torch.nn.Linear} layer is defined as \texttt{self.readout}. As with the latent layers, the learnable weight matrix \texttt{self.readout.weight} corresponding to \(\mathbf{W}^{\mathrm{out}}\) is initialized as Xavier uniform. This layer maps the topmost latent state $\mathbf{X}^{(L)}$ (with dimension $\texttt{dims[-1]} = d_L$) to the output prediction $\hat{\mathbf{Y}}$ (with dimension $\texttt{output\_dim} = d_{\rm out}$). In alignment with the supervised extension described, the layer implements $\hat{\mathbf{Y}} = \mathbf{X}^{(L)}\mathbf{W}^{\mathrm{out}\top}$ without a bias term.
\end{itemize}
The class includes two methods to facilitate the PCN's operation:
\begin{itemize}
    \item \textbf{\texttt{init\_latents(self, batch\_size, device)}}: This method initializes the latent variables as in lines 1--3 in Algorithm~3. For a given \texttt{batch\_size}, it creates a list of tensors $[\mathbf{X}^{(1)},\dots,\mathbf{X}^{(L)}]$ with values drawn independently from $\mathcal{N}(0,1)$, ensuring they are on the specified \texttt{device}. The flag \texttt{requires\_grad=False} ensures PyTorch's autograd engine does not compute or store gradients for these latent variables, as they are updated via the explicit inference rule instead of automatic differentiation of a loss function.
    \item \textbf{\texttt{compute\_errors(self, inputs\_latents)}}: This method calculates the prediction errors $\mathbf{E}^{(l)}$ and the gain-modulated errors $\mathbf{H}^{(l)}$ for all layers $0 \le l < L$. It takes a list $\texttt{inputs\_latents} = [\mathbf{X}^{(0)},\mathbf{X}^{(1)},\dots,\mathbf{X}^{(L)}]$ (containing the current input batch and all current latent states) as input. It iterates through each \texttt{PCNLayer} in \texttt{self.layers}, using the layer's \texttt{forward} method to get the prediction $\hat{\mathbf{X}}^{(l)}$ and pre-activation $\mathbf{A}^{(l)}$. It then computes $\mathbf{E}^{(l)} = \mathbf{X}^{(l)} - \hat{\mathbf{X}}^{(l)}$ and $\mathbf{H}^{(l)} = \mathbf{E}^{(l)} \odot f^{(l)'}(\mathbf{A}^{(l)})$. These computations are central to both the inference and learning phases, as seen in lines 6--9 and 21--24 of Algorithm~3. The method returns two lists: one containing all $\mathbf{E}^{(l)}$ and another containing all $\mathbf{H}^{(l)}$.
\end{itemize}

Together, these components and methods provide a modular PyTorch representation of the predictive coding network, suitable for the supervised learning task described.

\subsubsection{Training loop}
Finally, to show the operational aspects of the model, a minimal training loop is presented next. Focusing on essential update mechanisms, it differs slightly from the accompanying Python notebook, which also implements the energy-tracking features discussed earlier.

The function \texttt{train\_pcn} takes the PCN \texttt{model}, a PyTorch \texttt{data\_loader} for batching, the number of training \texttt{num\_epochs}, rates \texttt{eta\_infer} and \texttt{eta\_learn}, and the number of steps for inference (\texttt{T\_infer}) per sample and learning (\texttt{T\_learn}) per batch, along with the target \texttt{device}.

The training process begins by setting the \texttt{model} to training mode and moving it to the specified \texttt{device}. It then iterates for a given number of \texttt{num\_epochs}. Within each epoch, mini-batches of data (\texttt{x\_batch}, \texttt{y\_batch}) are processed. Input features \texttt{x\_batch} are flattened and, along with the one-hot encoded targets \texttt{y\_batch}, moved to the target \texttt{device}. The list \texttt{inputs\_latents} is initialized to hold the input batch $\mathbf{X}^{(0)}$ followed by the randomly initialized latent variables $\mathbf{X}^{(1)}, \dots, \mathbf{X}^{(L)}$ obtained from \texttt{model.init\_latents()}. Notably, a list named \texttt{weights} is created containing direct references to the model's weight tensors (\texttt{layer.W} for generative layers and \texttt{model.readout.weight} for the readout layer). Updates to elements of this \texttt{weights} list will therefore modify the model's parameters in place.

\begin{pythoncode}[title=Core training loop]
def train_pcn(model, data_loader, num_epochs, eta_infer, eta_learn,
				T_infer, T_learn, device='cuda'):
    model.to(device).train()

    for epoch in range(num_epochs):
        for x_batch, y_batch in data_loader:
            B = x_batch.size(0)
            d_0 = model.dims[0]
            x_batch = x_batch.view(B, d_0).to(device)
            y_batch = F.one_hot(y_batch, num_classes=model.readout.out_features) \
            			.float().to(device)
            inputs_latents = [x_batch] + model.init_latents(B, device)
            weights = [layer.W for layer in model.layers] + [model.readout.weight]

			# INFERENCE - T_infer steps
            with torch.no_grad(), autocast(device_type='cuda'):
                for t in range(1, T_infer + 1):
                    errors, gain_modulated_errors = model.compute_errors(inputs_latents)
                    y_hat           = model.readout(inputs_latents[-1])
                    eps_sup         = y_hat - y_batch
                    eps_L           = eps_sup @ weights[-1]
                    errors_extended = errors + [eps_L]
                    
                    # Latent gradients and updates
                    for l in range(1, model.L + 1):
                        grad_Xl = errors_extended[l] - \
                        				gain_modulated_errors[l-1] @ weights[l-1]
                        inputs_latents[l] -= eta_infer * grad_Xl

			# LEARNING - T_learn steps
            with torch.no_grad():          
                for t in range(T_infer + 1, T_learn + T_infer + 1):
                    errors, gain_modulated_errors = model.compute_errors(inputs_latents)
                    y_hat           = model.readout(inputs_latents[-1])
                    eps_sup         = y_hat - y_batch
                    
                    # Weight gradients and updates
                    for l in range(model.L):
                        grad_Wl = -(gain_modulated_errors[l].T @ inputs_latents[l+1]) / B
                        weights[l] -= eta_learn * grad_Wl
                    grad_Wout = eps_sup.T @ inputs_latents[-1] / B
                    weights[-1] -= eta_learn * grad_Wout
\end{pythoncode}

For each batch, the \textbf{inference phase} is executed for \texttt{T\_infer} steps. This phase operates under \texttt{torch.no\_grad()} context, as latent variable updates are performed manually according to the PCN rules, not via PyTorch's autograd. The \texttt{autocast} context manager is also used here, enabling mixed-precision computations on compatible CUDA hardware.
In each inference step:
\begin{enumerate}
    \item The \texttt{model.compute\_errors} method is called to calculate the current prediction errors $\mathbf{E}^{(l)}$ and gain-modulated errors $\mathbf{H}^{(l)}$ for the generative layers ($0 \le l < L$).
    \item The supervised prediction $\hat{\mathbf{Y}}$ and error $\mathbf{E}^{\mathrm{sup}}$ are computed.
    \item The error signal for the top latent layer, $\mathbf{E}^{(L)} = \mathbf{E}^{\mathrm{sup}}\mathbf{W}^{\mathrm{out}}$, is calculated. This notation aligns with how $\boldsymbol{\varepsilon}^{(L)}$ was defined for the supervised algorithm to use the general latent update rule. The full list of errors $[\mathbf{E}^{(0)}, \dots, \mathbf{E}^{(L)}]$ is assembled in \texttt{errors\_extended}.
    \item Each latent variable $\mathbf{X}^{(l)}$ (for $1 \le l \le L$) is updated using the formula $\mathbf{X}^{(l)} \gets \mathbf{X}^{(l)} - \eta_{\text{infer}}(\mathbf{E}^{(l)} - \mathbf{H}^{(l-1)}\mathbf{W}^{(l-1)})$, which corresponds to line 15--16 of Algorithm~3.
\end{enumerate}

Following inference, the \textbf{learning phase} adjusts the model weights for \texttt{T\_learn} steps, also under \texttt{torch.no\_grad()}. The latent variables \texttt{inputs\_latents} remain fixed at their values from the end of the inference phase. In each of these \texttt{T\_learn} steps:
\begin{enumerate}
    \item The generative errors $\mathbf{E}^{(l)}$, gain-modulated errors $\mathbf{H}^{(l)}$, and the supervised error $\mathbf{E}^{\mathrm{sup}}$ are recomputed. This is essential because the model weights (elements of the \texttt{weights} list) are updated within this loop, and thus the predictions and errors change accordingly.
    \item The generative weight matrices $\mathbf{W}^{(l)}$ (for $0 \le l < L$) are updated. The gradient $\mathbf{G}_{\mathbf{W}}^{(l)} = - \frac{1}{B} \mathbf{H}^{(l)\top} \mathbf{X}^{(l+1)}$ is computed, and the weights are adjusted: $\mathbf{W}^{(l)} \leftarrow \mathbf{W}^{(l)} - \eta_{\text{learn}}\mathbf{G}_{\mathbf{W}}^{(l)}$. This corresponds to lines 25--26 of Algorithm~3.
    \item Similarly, the readout weights $\mathbf{W}^{\mathrm{out}}$ are updated using their gradient $\mathbf{G}_{\mathbf{W}}^{\mathrm{out}} = \frac{1}{B} \mathbf{E}^{\mathrm{sup}\top}\mathbf{X}^{(L)}$, as per lines 30--31 of Algorithm~3.
\end{enumerate}

\subsection{Results and observations}
The model was trained in Google Colab on an NVIDIA L4 GPU, running the accompanying Python notebook~\cite{monadillo}. Training for just 4 epochs took 4 minutes. Yet the test performance scores in Table~\ref{tab:pcn-test-accuracies} are phenomenal. In fact, the top-1 accuracy (percentage correct) comfortably tops the leaderboard on \href{https://paperswithcode.com/sota/image-classification-on-cifar-10}{Papers With Code} at the time of writing (May~29, 2025)---the previous record of 99.5\% having been set by the vision transformer ViT-H/14 in 2020.

\begin{figure}[htbp]
    \centering
    \includegraphics[width=1.05\linewidth]{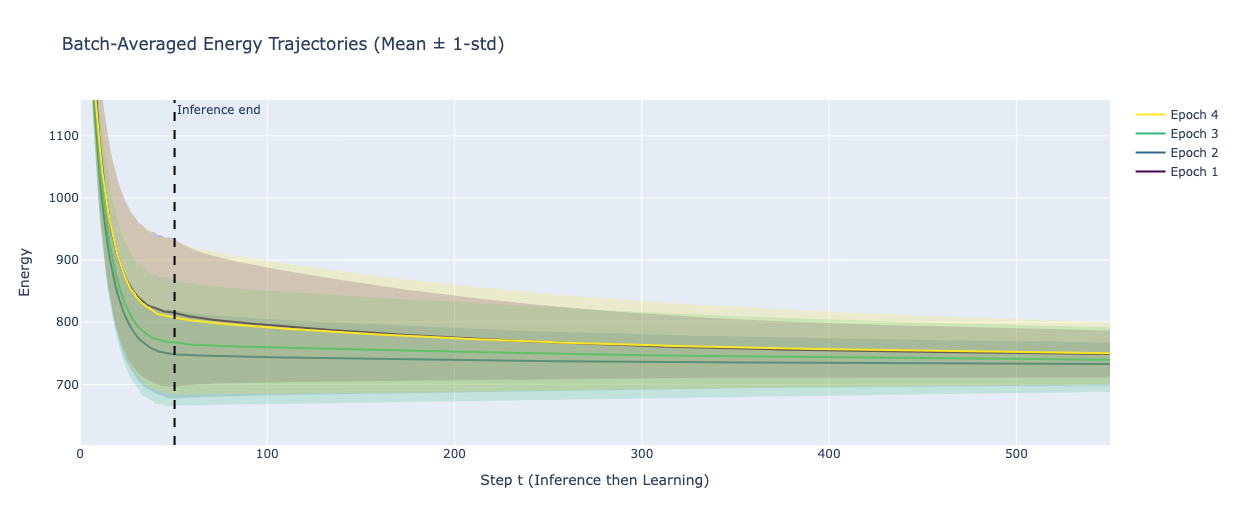}
    \includegraphics[width=1.0\linewidth]{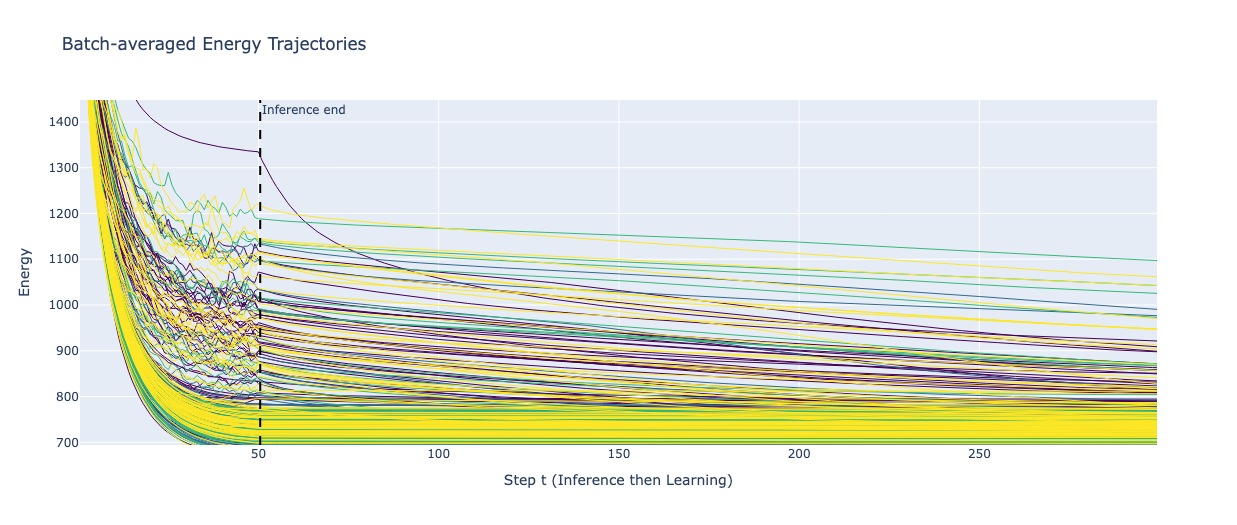}
    \caption{Partial zoom-in views of the batch-averaged energy trajectories recorded during training. The first plot shows the mean trajectories over all batches (equivalently, all samples) for each epoch, as well as the shaded areas of one standard deviation. The second plot shows the trajectories for individual batches. Coloring is by epoch. The full, interactive plots can be found in the Python notebook.}
    \label{fig:pcn_cifar10_experiment_plots}
\end{figure}

\vspace{1em}
\begin{table}[ht]
\centering
\begin{tabular}{l c}
\hline
Top-1 accuracy        & \scalebox{1}{99.92\%}         \\ 
Top-3 accuracy        & \scalebox{1}{99.99\%}         \\ 
\hline
\end{tabular}
\caption{PCN test accuracies after 4 epochs of training}
\label{tab:pcn-test-accuracies}
\end{table}
\noindent The batch-averaged energy trajectories recorded during training are depicted in Figure~\ref{fig:pcn_cifar10_experiment_plots}.

\vspace{1em}
\noindent{\bf Disclosure.}
The experiment was virtually one-shot. The rates were initially set to \(\eta_{\rm infer} = 0.1\) and \(\eta_{\rm learn} = 0.001\) for the very first run. Monitoring the batch-averaged energy trajectories during training, some batches displayed unstable inference while learning was generally slow. Consequently, the inference rate was cut in half to \(\eta_{\rm infer} = 0.05\) and the learning rate was quintupled to \(\eta_{\rm learn} = 0.005\) for the second run, which stabilized training and led to the test performance above on the first try. \emph{Neither the model architecture nor the hyperparameters were tuned for performance in any way or at any point.} In order to prevent data-leakage from the test set to validation, the experiment terminated here. (Observe that CIFAR-10 does not include a separate validation set.) The trained weights are publicly available~\cite{monadillo}.

\vspace{1em}
This little experiment demonstrates once more that predictive coding networks can be trained end-to-end on a basic vision task using  local, biologically plausible update rules. While generally not yet competitive with state-of-the-art deep learning methods in terms of scope and, depending on task, raw accuracy, PCNs offer an intriguing model of computation with sound theoretical and practical motivations.

In view of the disclosure above, the author has no idea whether a much smaller architecture (or different hyperparameters) might achieve similar performance. We leave it to the curious reader to tune the model and its training for that perfect \(100\%\) score. Happy hunting!~\smiley

\clearpage
\addcontentsline{toc}{section}{References}


\end{document}